\def\year{2021}\relax
\documentclass[letterpaper]{article} 
\usepackage{aaai21}  
\usepackage{times}  
\usepackage{helvet} 
\usepackage{courier}  
\usepackage[hyphens]{url}  
\usepackage{graphicx} 
\usepackage{natbib}  
\usepackage{caption} 
\usepackage{amsmath,amssymb}
\usepackage{tabularx,array,booktabs}
\usepackage{bm}
\usepackage{algorithm}
\usepackage{algorithmic}
\usepackage{tikz} 
\usepackage{hyperref} 
\usepackage{ifthen}

\usetikzlibrary{arrows,decorations.pathmorphing,backgrounds,fit,positioning,shapes.symbols,chains}
\tikzset{box1/.style={draw=black, thick, rectangle, minimum height=1.4cm, minimum width=3cm }}

\newcommand{\Section}[1]{\section{#1}\label{sec:#1}}
\newcommand{\Subsection}[1]{\subsection{#1}\label{sec:#1}}
\newcommand{\Subsubsection}[1]{\subsubsection{#1}\label{sec:#1}}

\title{AAMDRL: Augmented Asset Management with Deep Reinforcement Learning}

\author{
Eric Benhamou \textsuperscript{\rm 1,\rm 2}
David Saltiel \textsuperscript{\rm 1,\rm 3}, 
Sandrine Ungari \textsuperscript{\rm 4},
Abhishek Mukhopadhyay \textsuperscript{\rm 5},
Jamal Atif \textsuperscript{\rm 2} 
}
\affiliations{
	\textsuperscript{\rm 1} AI Square Connect, France, \texttt{\{eric.benhamou,david.saltiel\}@aisquareconnect.com} \\
	\textsuperscript{\rm 2}MILES, LAMSADE, Dauphine university, France, \texttt{eric.benhamou, jamal.atif@lamsade.dauphine.fr} \\\
	\textsuperscript{\rm 3} LISIC, ULCO, France, \texttt{david.saltiel@univ-littoral.fr} \\
	\textsuperscript{\rm 4} Societe Generale, Cross Asset Quantitative Research, UK,\\ 
	\textsuperscript{\rm 5} Societe Generale, Cross Asset Quantitative Research, France,\\ 
	\texttt{\{sandrine.ungari,abhishek.mukhopadhyay\}@sgcib.com}
}

\begin{document}
\maketitle
\thispagestyle{plain}
\pagestyle{plain}

\begin{abstract}
Can an agent learn efficiently in a noisy and self adapting environment with sequential, non-stationary and non-homogeneous observations? Through trading bots, we illustrate how Deep Reinforcement Learning (DRL) can tackle this challenge. Our contributions are threefold: (i) the use of contextual information also referred to as augmented state in DRL, (ii) the impact of a one period lag between observations and actions that is more realistic for an asset management environment, (iii) the implementation of a new repetitive train test method called walk forward analysis, similar in spirit to cross validation for time series. Although our experiment is on trading bots, it can easily be translated to other bot environments that operate in sequential environment with regime changes and noisy data. Our experiment for an augmented asset manager interested in finding the best portfolio for hedging strategies shows that AAMDRL achieves superior returns and lower risk.
\end{abstract}


\Section{Introduction}
Can a bot learn efficiently in a noisy and self adapting environment with sequential, non-stationary and non-homogeneous observations? By noisy and non homogeneous, we mean that data have different statistical properties across time. By sequential observations, we mean that chronological order matters and that observations are completely modified if we change their order. To answer this question, we use trading bots that offer a perfect example of noisy data, strongly sequential observations subject to change of regime. We aim at creating an augmented asset manager bot for the asset management industry.

Asset management is a well-suited industry to apply robotic machine learning: large amount of data available due to electronic tradings and strategic interest in bots as they do not give any hold on emotional and or behavioral bias largely described in \citet{Kahneman_2011} that can cause the asset manager ruin. However, machine learning is hardly used to make investment decision. Because of the complexity of the learning environment, asset managers are still largely relying on traditional methods, based on human decisions.

This is in sharp contrast with recent advances of deep reinforcement learning (DRL) on challenging tasks like game (Atari games from raw pixel inputs \citet{mnih-atari-2013, Mnih_2015}, Go \citet{Silver_2016}, StarCraft II \citet{Vinyals_2019}), but also more robotic learning like advanced locomotion and manipulation skills from raw sensory inputs (\citet{Levine_2015, Levine_2016} \citet{Schulman_2015,Schulman_2016,Schulman_2017}, \citet{Lillicrap_2015}), autonomous driving (\citet{Wang2018DeepRL}) and general bot learning (\citet{Gu_2017}).

We investigate if deep reinforcement learning can help creating an augmented asset manager when solving a classical portfolio allocation problem: finding hedging strategies to an existing portfolio, the MSCI World index in our example. The hedging strategies are different strategies operated by standard bots that have different logics and perform well in different market conditions. Knowing when to add and remove them and when to decrease or increase their asset under management is a fundamental but challenging question for an augmented asset manager.

\Subsection{Related works}
At first, reinforcement learning was not used in portfolio allocation. Initial works focused on trying to use deep networks to forecast next period prices, as presented in \citet{Freitas_2009}, \citet{Niaki2013}, \citet{Heaton_2017}. These models solved a supervised learning task akin to a regression, and tried to predict prices using past information only and compute portfolio allocations based on forecast. For asset managers, this initial usage of machine learning contains multiple problems. First, it does not ensure that the prediction is reliable in the near future: financial markets are well known to be non stationary and to  present regime changes as illustrated in \citet{Salhi_2016}, \citet{Dias_2015}, \citet{Zheng_2019}. Second, this approach does not address the question of finding the optimal portfolio based on some reward metrics. Third, it does not adapt to changing environment and does not easily incorporate transaction costs.

A second stream of research around deep reinforcement learning has emerged to address those points: \citet{Jiang_2016,Zhengyao_2017,Liang_2018,Yu_2019,Wang_2019,Liu_2020,Ye_2020,Li_2019,Xiong_2018}. The dynamic nature of reinforcement learning makes it an obvious candidate for changing environment  \citet{Jiang_2016,Zhengyao_2017,Liang_2018}, \citet{benhamou2020detecting}, \citet{Benhamou2020time}. Transaction costs can be easily included in rules \citet{Liang_2018,Yu_2019,Wang_2019,Liu_2020,Ye_2020,Yu_2019}. However, these works, except \citet{Ye_2020} rely on time series of open high low close prices, which are known to be very noisy. Secondly, they all assume an immediate action after observing prices which is quite different from reality as asset managers need a one day turnaround to manage new portfolio positions. Thirdly, they mostly rely on a single reward function and do not measure the impact of the reward function. Last but not least, they only do one train and test period, and never test for model stability.

More generally, DRL has recently been applied to other problems than portfolio allocation and direct trading strategies \citet{zhang2019deep}, \citet{huang2018financial},
\citet{thate2020application}, \citet{nan2020sentiment}, \citet{Wu_2020} or to the case of multi agents \citet{bao2019multiagent} or to optimal execution \citet{ning2018double}. 

\Subsection{Contributions}
Our contributions are threefold:
\begin{itemize}
\item \textbf{The addition of contextual information.} Using just past information is not sufficient for bot learning in a noisy and fast changing environment. The addition of contextual information improves results significantly. Technically, we create two sub-networks: one fed with direct observations (past prices and standard deviation) and another one with contextual information (level of risk aversion in financial markets, early warning indicators for future recession, corporate earnings...). 

\item \textbf{One day lag between price observation and action.} We assume that prices are observed at time $t$ but action only occurs at time $t+1$, to be consistent with reality. This one day lag makes the RL problem more realistic but also more challenging. 

\item \textbf{The walk-forward procedure.} Because of the non stationarity nature of time dependent data and especially financial data, it is crucial to test DRL models stability. We present a new methodology in DRL model evaluation referred to as walk forward analysis that iteratively trains and tests the model on extending data-set. This can be seen as the analogy of cross validation for time series. This allows to validate that selected hyper parameters work well over time and that resulting models are stable over time.
\end{itemize}

\Section{Background and mathematical formulation}\label{sec:Background}
In standard bot reinforcement learning, models are based on Markov Decision Process (MDP) as in \citet{SuttonBarto_2018}. MDP assumes that the bot knows all the states of the environment and has all the information to make the optimal decision in every state. The Markov property in addition implies that knowing the current state is sufficient.

Yet, the traditional MDP framework is inappropriate here: noise may arise in financial market data due to unpredictable external events. We prefer to use Partially Observable Markov Decision Process (POMDP) as presented initially in \citet{Astrom_1969}. In POMDP, only a subset of the information of a given state is available. The partially-informed agent cannot behave optimally. He uses a window of past observations to replace states as in a traditional MDP.

Mathematically, POMDP is a generalization of MDP. Recall that MDP assumes a 4-tuple $(\mathcal{S}, \mathcal{A}, \mathcal{P}, \mathcal{R})$ where $\mathcal{S}$ is the set of states, $\mathcal{A}$ is the set of actions, $\mathcal{P}$ is the state action to next state transition probability function $\mathcal{P} : \mathcal{S} \times \mathcal{A} \times \mathcal{S} \to \left[0, 1\right]$, and $\mathcal{R}$ is the immediate reward. 
The goal of the agent is to learn a policy that maps states to the optimal action $\mu : \mathcal{S} \to \mathcal{A}$ 
and that maximizes the expected discounted reward $\mathbb{E}[\sum_{t=0}^{\infty }\gamma ^{t}R_t]$. 
POMPD adds two more variables in the tuple, $\mathcal{O}$ and $\mathcal{Z}$ where $\mathcal{O}$ is the set of observations and $\mathcal{Z}$ is the observation transition function $\mathcal{Z}: \mathcal{S} \times \mathcal{A} \times \mathcal{O} \to [0,1]$. 
At each time, the agent is asked to take an action $a_t \in \mathcal{A}$ in a particular environment state $s_t \in \mathcal{S}$, that is followed by the next state $s_{t+1}$ with transition probability $\mathcal{P}(s_{t+1}| s_t, a_t )$. The next state $s_{t+1}$ is not observed by the agent. It rather receives an observation $o_{t+1} \in \mathcal{O}$ on the state $s_{t+1}$ with probability $Z(o_{t+1}| s_{t+1}, a_t )$. 

From a practical standpoint, the general RL setting is modified by taking a pseudo state formed with a set of past observations $(o_{t-n}, o_{t-n-1}, \ldots, o_{t-1}, o_t)$. In practice to avoid large dimension and the curse of dimension, it is useful to reduce this set and take only a subset of these past observations with $j< n$ past observations, such that $0<i_1< \ldots < i_j$ and $i_{k, 1\leq k \leq j} \in \mathbb{N}$ is an integer. The set $\delta_1 = (0,i_1, \ldots, i_j)$ is called the observation lags. In our experiment we typically use lag periods like (0, 1, 2, 3, 4, 20, 60) for daily data, where the tuple $(0,1,2,3,4)$ is indeed the last week observation, $20$ is for the one-month ago observation (as there is approximately 20 business days in a month) and 60 the three-month ago observation.

\Subsection{Observations}
\Subsubsection{Regular observations}
There are two types of observations: regular and contextual information. Regular observations are data directly linked to the problem to solve. For a standard bot, these are observations from its environment like position of the arm, degree, etc. In the case of a trading bot, regular observations are past prices observed over a lag period $\delta = (0<i_1< \ldots < i_j)$. To re-normalize data, we rather use past returns computed as  $r_t^k = \frac{p^k_t}{p^k_{t-1} } -1$ where $p^k_t$ is the price at time $t$ of the asset $k$. For a financial asset $k$, to give information about regime changes, our trading bot receives also empirical standard deviation computed over a sliding estimation window denoted by $d$ as follows $\sigma^k_t  = \sqrt{ \frac{1}{d} \sum_{u =t-d+1}^t \left( r_u^k - \mu \right)^2 }$, where the empirical mean $\mu^k$ is computed as $\mu^k = \frac{1}{d} \sum_{u =t-d+1}^t r
^k_u$. Hence our regular observations is a three dimensional tensor represented as follows:

\begin{center}
\begin{tikzpicture}
[
comment/.style={rectangle, inner sep= 0pt, text width=3cm, node distance=0.25cm, font=\scriptsize\sffamily},
comment2/.style={rectangle, inner sep= 0pt, text width=3cm, node distance=0.25cm, font=\sffamily}
]
\node[box1, fill=red, fill opacity=0.1, color = blue!80] (c2) at (0,0) {};
\node[box1, fill=blue, fill opacity=0.1, color = blue!60] (c1) at (0.3,0.3) { };
\node [comment2, right= -2.5 cm of c2, color = black!80] {Returns $A^1_t$ };
\node [comment2, right= -2.5 cm of c1, color =  black!60] {Volatility $A^2_t$ };
\draw[](0,0){};
\end{tikzpicture}  \\
\vspace{0.3cm}
with \,\, $A^1_t =  \left( \!
\begin{array}{c   }
r^1_{t-i_j} \,\,	... \,\, r^1_t \\
... \,\,... \,\, ...\\
r^m_{t-i_j} \,\,.... \,\, r^m_t
\end{array} \! \right)\! ,  \,\,
 A^2_t =  \left( \!
\begin{array}{c  }
\sigma^1_{t-i_j} 	\,\,	... \,\, \sigma^1_t\\
... \,\,... \,\, ...\\
\sigma^m_{t-i_j} \,\,.... \,\, \sigma^m_t
\end{array} \! \right)$
\end{center}
\vspace{0.3cm}

This setting with two layers (past returns and past volatilities) is quite different from the one presented in \citet{Jiang_2016,Zhengyao_2017,Liang_2018} that uses different layers representing open, high, low and close prices. There are various remarks to be made. First, high low information does not make sense for portfolio strategies that are only evaluated daily, which is the case of all the funds. Secondly, open high low  prices tend to be highly correlated creating some noise in the inputs. Third, the concept of volatility is crucial to detect regime change and is surprisingly absent from these works as well as from other works like \citet{Yu_2019,Wang_2019,Liu_2020,Ye_2020,Li_2019,Xiong_2018}.

\Subsubsection{Context observation}\label{Context}
Contextual observations are additional information that provides intuition about current context. For our asset manager bot, they are other financial data not directly linked to its portfolio assumed to have some predictive power for portfolio assets. In the case of a financial portfolio, context information is typically modelled by a large range of features :
\begin{itemize}
    \item the level of risk aversion in financial markets, or market sentiment, measured as an indicator varying between 0 for maximum risk aversion and 1 for maximum risk appetite,
    \item the bond/equity historical correlation, a classical ex-post measure of the diversification benefits of a duration hedge, measured on a 1-month, 3-month and 1-year rolling window,
    \item The credit spreads of global corporate - investment grade, high yield, in Europe and in the US - known to be an early indicator of potential economic tensions, 
    \item The equity implied volatility, a measure if the 'fear factor' in financial market,
    \item The spread between the yield of Italian government bonds and the German government bond, a measure of potential tensions in the European Union,
    \item The US Treasury slope, a classical early indicator for US recession,
    \item And some more financial variables, often used as a gauge for global trade and activity: the dollar, the level of rates in the US, the estimated earnings per shares (EPS).
\end{itemize}

On top of these observations, we also include the maximum and minimum portfolio strategies return and the maximum portfolio strategies volatility. The latter information is like for regular observations motivated by the stylized fact that standard deviations are useful features to detect crisis. 

\begin{figure}[htbp]
\centering
\includegraphics[width=\linewidth]{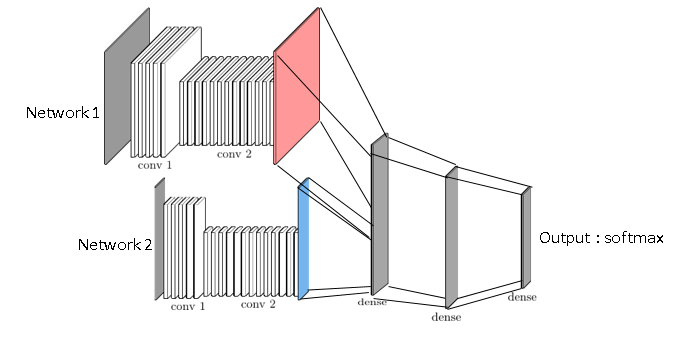}
\caption{Network architecture}\label{fig:best_network}
\end{figure}

Contextual observations are stored in a 2D matrix denoted by $C_t$ with stacked past $p$ individual contextual observations. The contextual state writes as $
C_t =  \left( \!\!
\begin{array}{c  }
c^1_t 	\,\,	... \,\, c^1_{t-i_k}\\
... \,\,... \,\, ...\\
c^p_t 	\,\,.... \,\, c^p_{t-i_k}
\end{array} \!\!\! \right)$. The matrix nature of contextual states $C_t$ implies in particular that we will use 1D convolutions should we use convolutional layers. All in all, observations that are augmented observations, write as $O_t =[ A_t, C_t]$, with $A_t=[A^1_t, A^2_t]$ that will feed the two sub-networks of our global network as presented in figure \ref{fig:best_network}.

\Subsection{Action}
In our deep reinforcement learning the augmented asset manager trading bot needs to decide at each period in which hedging strategy it invests. The augmented asset manager can invest in $l$ strategies that can be simple strategies or strategies that are also done by asset management bots. To cope with reality, the bot will only be able to act after one period. This is because asset managers have a one day turn around to change their positions. We will see on experiments that this one day turnaround lag makes a big difference in results. As it has access to $l$ potential hedging strategies, the output is a $l$  dimension vector that provides how much it invests in each hedging strategy. For our deep network, this means that the last layer is a softmax layer to ensure that portfolio weights are between $0$ and $100\%$ and sum to $1$.

\Subsection{Reward}
There are multiple choices for our reward. A straightforward reward function is to compute the final net performance of the combination of our portfolio computed as the value of our portfolio at the last train date $t_T$ over the initial value of the portfolio $t_0$ minus one: $P_{t_T} / P_{t_0} -1$. Another natural reward function is to compute the Sortino ratio, that is a variation of Sharpe ratio where risk is computed by the downside standard deviation (instead of regular standard deviation) whose definition is to compute the standard deviation only on negative daily returns denoted by $(\tilde{r}_t)_{t=0..T}$ . Hence the downside standard deviation is computed by $\sqrt{250} \times  \text{StdDev}[ (\tilde{r}_t)_{t=0..T}]$.

\Subsection{Adversarial Policy Gradient}
A policy is a mapping from the observation space to the action space, $\pi:\mathcal{O}\rightarrow\mathcal{A}$. 
To achieve this, a policy is specified by a deep network with a set of parameters $\vec \theta$. The action is a vector function of the observation given the parameters: $\vec a_t = \pi_{\vec \theta}(\bm o_t)$.
The performance metric of $\pi_{\vec \theta}$ for time interval $[0,t]$ is defined as the corresponding total reward function of the interval $
	J_{[0,t]}(\pi_{\vec \theta}) = R\left( \vec o_1,\pi_{\vec \theta}(o_1),\cdots,
		\vec o_{t},\pi_{\vec \theta}(o_{t}),\vec o_{t+1} \right)
	\label{eq:policy_value}$.
After random initialization, the parameters are continuously updated along the gradient direction with a learning rate $\lambda$:
$\vec\theta \longrightarrow \vec\theta + \lambda\nabla_{\vec\theta}J_{[0,t]}(\pi_{\vec \theta})$. The gradient ascent optimization is done with standard Adam (short term for Adaptive Moment Estimation) optimizer to have the benefit of adaptive gradient descent with root mean square propagation \citet{kingma2014method}. The whole process is summarized in algorithm \ref{alg1}, called adversarial policy gradient as we introduce randomisation both in the observations and the action (to have standard exploration exploitation). This two steps randomization ensures more robust training as we will see in the experiments. Noise in observations has already been suggested to improve training in \citet{Liang_2018}.

\begin{algorithm}[htbp]
    \caption{Adversarial Policy Gradient}
    \label{alg1}
\begin{algorithmic}[1]
    \STATE Input: initial policy parameters $\theta$, empty replay buffer $\mathcal{D}$
\REPEAT
    \STATE reset replay buffer
    \WHILE{not terminal}
        \STATE Observe observation $o$ and select action $a = \pi_{\theta}(o)$ with probability $p$ and random action with probability $1-p$, 
        \STATE Execute $a$ in the environment    
        \STATE Observe next observation $o'$, reward $r$, and done signal $d$ to indicate whether $o'$ is terminal
        \STATE apply noise to next observation $o'$
        \STATE store $(o,a,o')$ in replay buffer $\mathcal{D}$
        \IF{Terminal}
            \FOR{however many updates in $\mathcal{D}$}
                \STATE compute final reward $R$
            \ENDFOR
            \STATE update network parameter with Adam gradient ascent
                $\vec\theta \longrightarrow \vec\theta + \lambda\nabla_{\vec\theta}J_{[0,t]}(\pi_{\vec \theta})$
        \ENDIF
    \ENDWHILE
\UNTIL{convergence}
\end{algorithmic}
\end{algorithm}

\Subsection{Walk forward analysis}
In machine learning, the standard approach is to do $k$-fold cross validation as shown in figure \ref{fig:cross_val}. White rectangles represent training periods while grey rectangles testing periods. This approach breaks the chronology of data and potentially uses past data in the test set. Rather, we can take sliding test set and take past data as training data. We can either take the training data set with a fixed starting point and grow the training data set by adding more and more data, which is what we call extending walk forward as shown in figure \ref{fig:extending} or take always the same amount of data and slide the training data (figure \ref{fig:sliding}), hence the name of sliding walk forward. Extending walk forward tends to more stable models as we add incrementally new data, at each new training step, and share all past data. The negative effect of this is to adapt slowly to new information. On the opposite, sliding walk forward leads to more rapidly changing models as we progressively drop old data and hence give more weight to more recent data. To our experience, because we do not have so much data to train our DRL model, it is better to use extending walk forward. Last but not least, as the test set is always after the train set, walk forward analysis gives less steps compared to cross validation. In practice for our data set, we train our models from 2000 to end of 2006 (to have at least seven years of data) and use an extending test period of one year.

\def\a{-1}
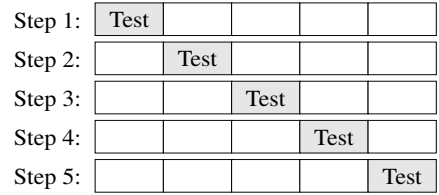
\begin{figure} [htbp]
\centering
\resizebox{0.7 \linewidth} {2.6cm} {
\begin{tikzpicture}[]
\foreach \i [count=\j from 0] in {white,white,white,white,gray!20} 
 \draw(\a+\j,0) rectangle (\a+\j+1,0.5)[fill=\i] ;
\foreach \i [count=\j from 0] in {white,white,white,gray!20,white} 
 \draw(\a+\j,0.6) rectangle (\a+\j+1,1.1)[fill=\i] ;
\foreach \i [count=\j from 0] in {white,white,gray!20,white,white} 
 \draw(\a+\j,1.2) rectangle (\a+\j+1,1.7)[fill=\i] ;
\foreach \i [count=\j from 0] in {white,gray!20,white,white,white} 
 \draw(\a+\j,1.8) rectangle (\a+\j+1,2.3)[fill=\i] ;
\foreach \i [count=\j from 0] in {gray!20,white,white,white,white} 
 \draw(\a+\j,2.4) rectangle (\a+\j+1,2.9)[fill=\i] ;
\foreach \i [count=\j from 0] in { , , , , } 
\draw(\a+4.5-\j,\j*0.6) node[above]{Test};
\foreach \i [count=\j from 1] in { , , , , } 
\draw(\a-0.7,2.92-\j*0.6) node[above]{Step \j:};
\end{tikzpicture}
}
\caption{k-fold cross validation} \label{fig:cross_val}
\end{figure}
\def\b{-1}

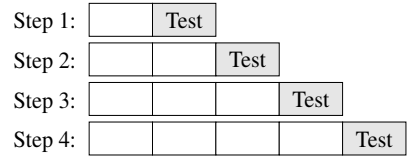
\begin{figure}[htbp]
\centering
\resizebox{0.65 \linewidth} {2.1cm} {
\begin{tikzpicture}[]
\foreach \i [count=\j from 0] in {white,white,white,white,gray!20} 
 \draw(\b+\j,0.6) rectangle (\b+\j+1,0.5+0.6)[fill=\i] ;
\foreach \i [count=\j from 0] in {white,white,white,gray!20} 
 \draw(\b+\j,0.6+0.6) rectangle (\b+\j+1,1.1+0.6)[fill=\i] ;
\foreach \i [count=\j from 0] in {white,white,gray!20} 
 \draw(\b+\j,1.2+0.6) rectangle (\b+\j+1,1.7+0.6)[fill=\i] ;
\foreach \i [count=\j from 0] in {white,gray!20}  
\draw(\b+\j,1.8+0.6) rectangle (\b+\j+1,2.3+0.6)[fill=\i] ;
\foreach \i [count=\j from 0] in { , , ,} 
\draw(\b+4.5- \j,\j*0.6+0.6) node[above]{Test};
\foreach \i [count=\j from 1] in { , , , } 
\draw(\a-0.7,2.92-\j*0.6) node[above]{Step \j:};
\end{tikzpicture}
}
\caption{Extending Walk Forward} \label{fig:extending}
\end{figure}

\def\c{-1}
\begin{figure}[htbp]
\centering
\resizebox{0.7 \linewidth} {2cm} {
\begin{tikzpicture}[]
\foreach \i [count=\j from 4] in {white,gray!20} 
 \draw(\c+\b+\j,0.6) rectangle (\c+\b+\j+1,0.5+0.6)[fill=\i] ;
\foreach \i [count=\j from 3] in {white,gray!20} 
 \draw(\c+\b+\j,0.6+0.6) rectangle (\c+\b+\j+1,1.1+0.6)[fill=\i] ;
\foreach \i [count=\j from 2] in {white,gray!20} 
 \draw(\c+\b+\j,1.2+0.6) rectangle (\c+\b+\j+1,1.7+0.6)[fill=\i] ;
\foreach \i [count=\j from 1] in {white,gray!20} 
 \draw(\c+\b+\j,1.8+0.6) rectangle (\c+\b+\j+1,2.3+0.6)[fill=\i] ;
\foreach \i [count=\j from 0] in { , , ,} 
\draw(\c+\b+5.5- \j,\j*0.6+0.6) node[above]{Test};
\foreach \i [count=\j from 1] in { , , , } 
\draw(\a-0.7,2.92-\j*0.6) node[above]{Step \j:};
\end{tikzpicture}
}
\caption{Sliding Walk Forward} \label{fig:sliding}
\end{figure}
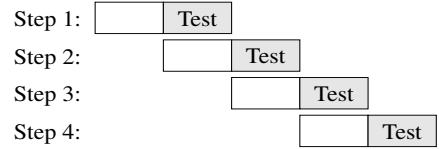

\Section{Experiments}

\Subsection{Goal of the experiment}
We are interested in finding a hedging strategy for a risky asset. The experiment is using daily data from 01/05/2000 to 19/06/2020. The risky asset is the MSCI world index (see  \href{https://www.msci.com/}{https://www.msci.com/} - data source: SG CIB). We choose the MSCI world index because it is a good proxy for a wide range of asset manager portfolios. The hedging strategies are 4 SG CIB proprietary systematic strategies (see \href{https://sgi.sgmarkets.com/}{https://sgi.sgmarkets.com/}  - data source: SG CIB), computed and executed by trading bots and further described below. As we use extending walk forward traning, train set is initially from 2000 to 2006 with a test set in 2007, then training set is from 2000 to 2007, with a test set in 2008, and etc up to the last training set which is from 2000 to 2019 with a test set in 2020.

\Subsection{Data-set description}\label{Data-set}
Systematic strategies are asset management bots that invest in financial markets according to  adaptive and pre-defined trading rules.  Here, we use 4 SG CIB proprietary 'hedging strategies', that tend to perform when stock markets are down:

\begin{itemize}
\item Directional hedges - react to small negative return in equities,
\item Gap risk hedges - perform well in sudden market crashes,
\item Proxy hedges - tend to perform in some market configurations, like for example when highly indebted stocks under-perform other stocks,
\item Duration hedges - invest in bond market, a classical diversifier to equity risk in finance. 
\end{itemize}

The underlying financial instruments vary from put options, listed futures, single stocks, to government bonds. Some of those strategies are akin to an insurance contract and bear a negative cost over the long run. The challenge consists in balancing cost versus benefits. In practice, asset managers have to decide how much of these hedging strategies are needed on top of an existing portfolio to achieve a better risk reward. The decision making process is often based on contextual information, such as the economic and geopolitical environment, the level of risk aversion among investors and other correlation regimes.  A cross validation step selects the most relevant features contextual information. In the present case, the first three features are selected. The rebalancing of strategies in the portfolio comes with transaction costs, that can be quite high since hedges use options. Transactions costs are like frictions in physical systems. They are taken into account dynamically to penalise solutions with a high turnover rate.

\Subsection{Evaluation metrics}
Asset managers use a wide range of metrics to gauge the success of their investment decision. For a thorough review of those metrics, see for example \citet{Cogneau_2009}. To keep things simple, we use the following metrics:
\begin{itemize}
\item annualized return defined as the average annualized compounded return,
\item annualized daily based Sharpe ratio defined as the ratio of the annualized return over the annualized daily based volatility $\mu / \sigma$,
\item Sortino ratio computed as the ratio of the annualized return overt the downside standard deviation,
\item maximum drawdown denoted by max DD in table \ref{tab:Model comparison}.
\end{itemize}

Let $P_T$ be the final value of the portfolio at time $T$ and $P_0$ its initial value at time $t=0$. Let $\tau$ be the year fraction of the final time $T$. The annualized return is defined as $\mu=(P_T / P_0)^{1/\tau} -1$. The maximum drawdown is computed as the maximum of all daily drawdowns. The daily drawdown is computed as the ratio of the difference between the running maximum of the portfolio value ($RM_T = \max_{t=0..T}(P_t)$ ) and the portfolio value over the running maximum of the portfolio value. Hence $DD_T = (RM_T - P_T) / RM_T$ and $MDD_T = \max_{t=0..T}(DD_t)$. 

\Subsection{Baseline}
\Subsubsection{Pure risky asset}
This first evaluation is to compare our portfolio composed only of the risky asset (in our case, the MSCI world index) with the one augmented by the trading bot and composed of the risky asset and the hedging overlay. If our bot is successful in identifying good hedging strategies, it should improve the overall portfolio and have a better performance than the risky asset.

\Subsubsection{Markowitz theory}
The standard approach for portfolio allocation in finance is the Markowitz model (\citet{Markowitz_1952}). It computes the portfolio with minimum variance given an expected return which is taken in our experiment to be the average return of the hedging strategies over the last year. The intuition in Markowitz (or mean-variance portfolio) theory is that an investor wants to have the lowest risk for a given return. In practice, we solve a quadratic program that finds the minimum portfolio variance under the constraint that the expected return is greater or equal to the minimum return. In our baseline, Markowitz portfolio is recomputed every 6 months to have something dynamic to cope with regime changes.

\Subsubsection{Follow the winner}
This is a simple strategy that consists in selecting the hedging strategy that was the best performer in the past year. If there is some persistence over time of the hedging strategies' performance, this simple methodology should work well. It replicates standard investors behavior that tend to select strategies that performed well in the past.

\Subsubsection{Follow the loser}
Follow the loser is the opposite of follow the winner. It assumes that there is some mean reversion in strategies' performance, meaning that strategies tend to perform equally well on long term and mean revert around their trend. Hence if a strategy did not perform well in the past, and if there is mean reversion, there is a lot of chance that this strategy will recover with its pairs.

\Subsection{Results and discussion}\label{sec:Results}
We compare the performance of the following 5 models: DRL model based on convolutional networks with contextual states (Sentiment indicator, 6 month correlation between equity and bonds and credit main index), same DRL model without contextual states, follow the winner, follow the loser and Markowitz portfolio. The resulting graphics are displayed in figure \ref{fig:netprofit_performance} with the risky asset position alone in blue and the models in orange. Out of these 5 models, only DRL and Follow the winner are able to provide significant net performance increase thanks to an efficient hedging strategy over the 2007 to 2020 period. The DRL model is in addition able to better adapt to the Covid crisis and to have better efficiency in net return but also Sharpe and Sortino ratios over 3 and 5 years as shown in table \ref{tab:Model comparison}. In terms of the smallest maximum drawdown, the follow the loser model is able to significantly reduce maximum drawdown but at the price of a lower return, Sharpe and Sortino ratios. Removing contextual information deteriorates model performances significantly and is illustrated by the difference in term of return, Sharpe, Sortino ratio and maximum drawdown between the DRL and the DRL no context model. Last but not least, Markowitz model is not able to adapt to the new regime change of 2015 onwards despite its good performance from 2007 to 2015. It is the worst performer over the last 3 and 5 years because of this lack of adaptation. For all models, we use the walk forward analysis as described in the corresponding section. Hence, we start training the models from 2000 to end of 2006 and use the best model on the test set in 2007. We then train the model from 2000 to end of 2007 and use the best model on the test set in 2008. In total, we do 14 training (from 2007 to 2020). This process ensures that we detect models that are unstable over time and is similar in spirit to delayed online training.  

\Subsection{Impact of context}
In table \ref{tab:model choice}, we provide a list of 32 models based on the following choices: network architecture (LSTM or CNN), adversarial training with noise in data or not, use of contextual states, and reward function (net profit and Sortino), use of day lag between observations and actions. We see that the best DRL model with the day-lag turnover constraint is the one using convolutional networks, adversarial training, contextual states and net profit reward function. These 4 parameters are meaningful for our DRL model and change model performance substantially as illustrated by the table. We also compare the same model with and without contextual state and see in table \ref{tab:Impact of contextual state} that the use of contextual state improves model performance substantially. This is quite intuitive as we provide more meaningful data to the model.

\Subsection{Impact of one day lag}
Reminding the fact that asset managers cannot immediately change their position at the close of the financial markets, modeling the one day lag turnover to account is also significant as shown in table \ref{tab:Impact of daylag}. It is not surprising that a delayed action after observation makes the learning process more challenging for the DRL agent as influence of variables tends to decrease with time. Surprisingly, this salient modeling characteristic is ignored in existing literature \citet{Zhengyao_2017,Liang_2018,Yu_2019,Wang_2019,Liu_2020,Ye_2020,Li_2019}.

\begin{table}[htbp]
  \centering
  \caption{Model comparison based on reward function, network (CNN or LSTM units) adversarial training (noise in data) and use of contextual state}
  \label{tab:model choice}%
\resizebox{\linewidth} {!} {
    \begin{tabular}{|lccccr|}
    \toprule
    reward	& network & adversarial 	& contextual 	& performance       & performance  \\
    			& 		 & training 		& states 		&  with 			& 			with \\
    			&		& 				&			& 1 day lag   		& 0 day lag \\
    \midrule
    Net\_Profit & CNN    & Yes    & Yes    & 81.8\%  & 123.8\% \\
    Net\_Profit & CNN    & No     & Yes    & 75.2\%  & 112.3\% \\
    Net\_Profit & LSTM   & Yes    & Yes    & 65.9\%  & 98.8\% \\
    Net\_Profit & LSTM   & No     & Yes    & 64.5\%  & 98.5\% \\
    Sortino & LSTM   & No     & Yes        & 61.8\%  & 87.4\% \\
    Net\_Profit & LSTM   & No     & No     & 56.6\%  & 59.8\% \\
    Sortino & LSTM   & No     & No         & 48.5\%  & 51.4\% \\
    Net\_Profit & LSTM   & Yes    & No     & 47.5\%  & 50.8\% \\
    Sortino & LSTM   & Yes    & Yes        & 29.6\%   & 47.6\% \\
    Sortino & LSTM   & Yes    & No         & 28.4\%  & 47.0\% \\
    Sortino & CNN    & No     & Yes        & 26.5\%  & 45.3\% \\
    Sortino & CNN    & Yes    & Yes        & 26.3\%  & 29.3\% \\
    Sortino & CNN    & Yes    & No         & -16.7\% & 16.9\% \\
    Net\_Profit & CNN    & Yes    & No     & -29.5\% & 13.9\% \\
    Sortino & CNN    & No     & No         & -45.0\% & 10.6\% \\
    Net\_Profit & CNN    & No     & No     & -47.7\% & 8.6\% \\
    \bottomrule
    \end{tabular}%
    }
\end{table}%

\begin{table}[htbp]
  \centering
  \caption{Impact of day lag}
\resizebox{\linewidth} {!} {
    \begin{tabular}{|llclr|}
    \toprule
    reward	& network & adversarial 	& contextual 	& day  lag 	impact \\
    			&   		& training 		& states 		&	\\
    \midrule
    Net\_Profit & CNN    & Yes    & Yes    & -42.0\% \\
    Net\_Profit & CNN    & No     & Yes    & -37.2\% \\
    Net\_Profit & LSTM   & Yes    & Yes    & -32.9\% \\
    Net\_Profit & LSTM   & No     & Yes    & -34.0\% \\
    Sortino & LSTM   & No     & Yes    & -25.6\% \\
    Net\_Profit & LSTM   & No     & No     & -3.2\% \\
    Sortino & LSTM   & No     & No     & -2.9\% \\
    Net\_Profit & LSTM   & Yes    & No     & -3.3\% \\
    Sortino & LSTM   & Yes    & Yes    & -18.0\% \\
    Sortino & LSTM   & Yes    & No     & -18.7\% \\
    Sortino & CNN    & No     & Yes    & -18.8\% \\
    Sortino & CNN    & Yes    & Yes    & -3.0\% \\
    Sortino & CNN    & Yes    & No     & -33.6\% \\
    Net\_Profit & CNN    & Yes    & No     & -43.4\% \\
    Sortino & CNN    & No     & No     & -55.6\% \\
    Net\_Profit & CNN    & No     & No     & -56.3\% \\
    \bottomrule
    \end{tabular}%
}
  \label{tab:Impact of daylag}%
\end{table}%

\begin{table}[htbp]
  \centering
   \caption{Impact of contextual state}
\resizebox{0.8 \linewidth} {!} {
    \begin{tabular}{|llcr|}
    \toprule
    reward	& network & adversarial 	&  contextual states  \\
    			& 		 & training 		& {impact} 		\\
    \midrule
    Net\_Profit & CNN    & Yes    & 111.4\% \\
    Net\_Profit & CNN    & No     & 122.9\% \\
    Net\_Profit & LSTM   & Yes    & 18.5\% \\
    Net\_Profit & LSTM   & No     & 7.9\% \\
    Sortino & LSTM   & No     & 13.3\% \\
    Sortino & LSTM   & Yes    & 1.2\% \\
    Sortino & CNN    & No     & 71.5\% \\
    Sortino & CNN    & Yes    & 43.0\% \\
    \bottomrule
    \end{tabular}%
}
  \label{tab:Impact of contextual state}%
\end{table}%

\def\gheight{3cm}
\def\gwidth{0.49}
\begin{figure}[htbp]
\centering
\includegraphics[width=\gwidth \linewidth, height=\gheight]{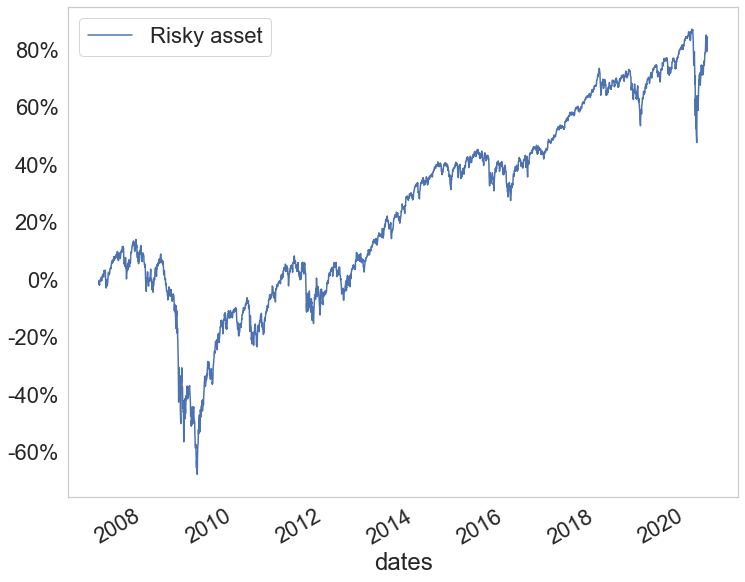}
\includegraphics[width=\gwidth \linewidth, height=\gheight]{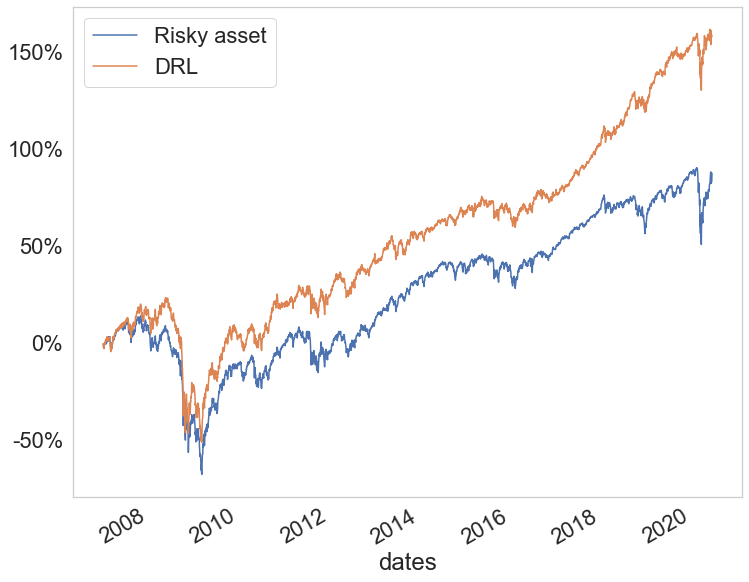}
\includegraphics[width=\gwidth \linewidth, height=\gheight]{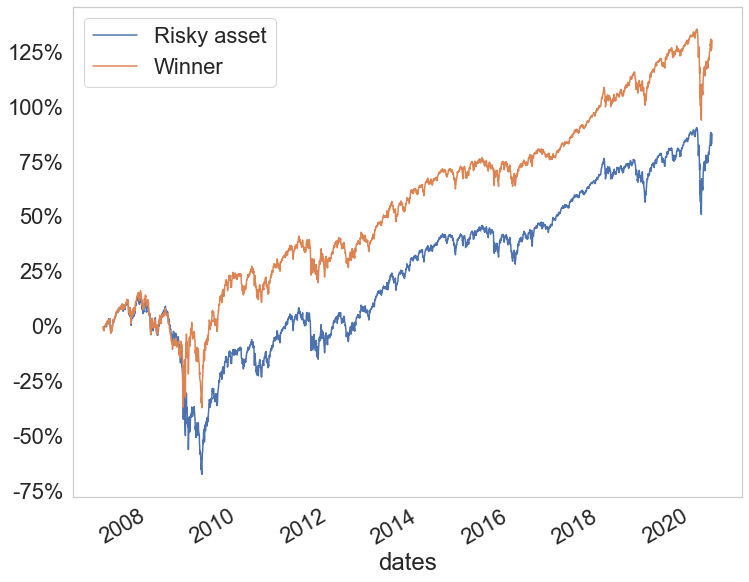}
\includegraphics[width=\gwidth \linewidth, height=\gheight]{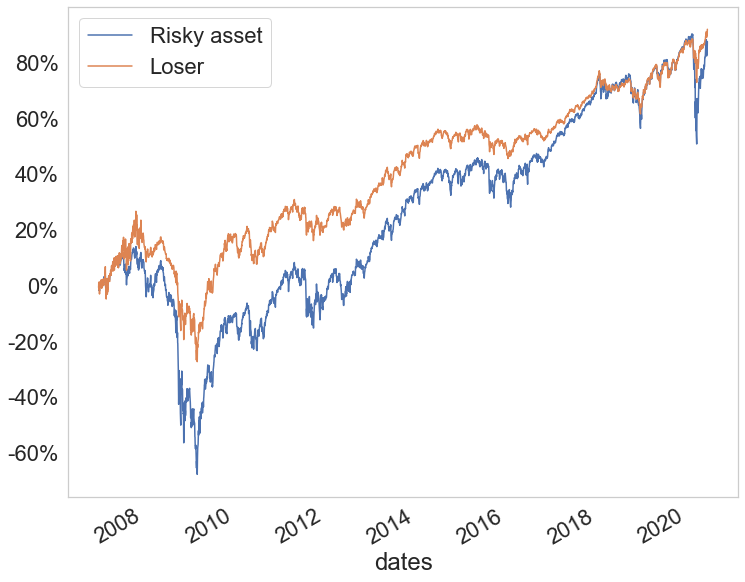}
\includegraphics[width=\gwidth \linewidth, height=\gheight]{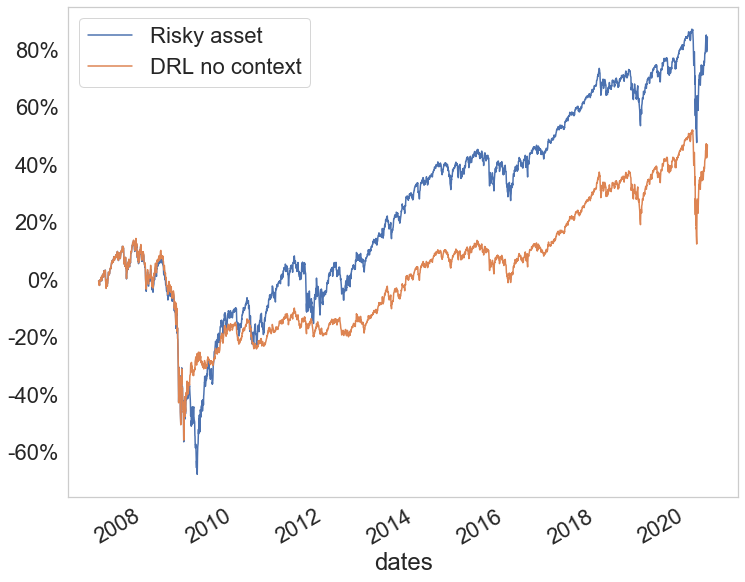}
\includegraphics[width=\gwidth  \linewidth, height=\gheight]{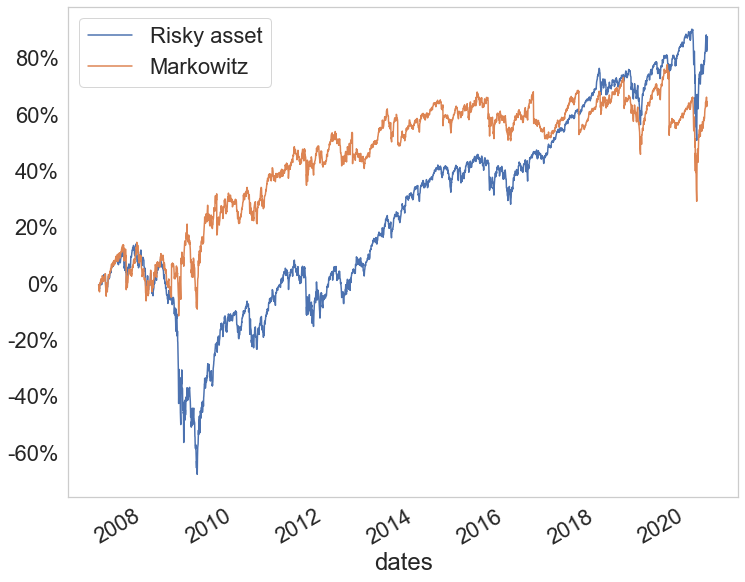}
\caption{from left to right and from top to bottom, performance for risky asset, DRL, follow the winner, follow the loser, DRL without context and Markowitz models. Most models are not able to continuously adapt to regime changes and consequently under-perform compared to the standalone risky asset position on a long period like 2007 to 2020.}\label{fig:netprofit_performance}
\end{figure}

\begin{table}[htbp]
  \centering
  \caption{Models comparison over 3 and 5 years}\label{tab:Model comparison}
\resizebox{0.95 \linewidth} {!} {
    \begin{tabular}{|l|rrrr|rrrr|}
    \toprule
	 &        & 3 Years &        &        \\
    \toprule   
        & \multicolumn{1}{l}{return } & \multicolumn{1}{l}{Sortino} & \multicolumn{1}{l}{Sharpe} & \multicolumn{1}{l|}{max DD} 		\\
    Risky asset & 10.27\% &       0.34  &      0.38  & -    0.34  		\\
    DRL  & \textbf{22.45\%} & \textbf{1.18} & \textbf{1.17} & -0.27 	\\
    Winner & 13.19\% & 0.66  & 0.72  & -0.35  						 \\
    Loser  & 9.30\% & 0.89  & 0.89  & \textbf{-0.15}  				\\
    DRL no context & 8.11\% & 0.42  & 0.47  & -0.34  				 \\
    Markowitz & -0.31\% & -0.01  & -0.01  & -0.41  					 \\
    \bottomrule
    \toprule
& 5 Years &        &  \\
    \toprule   
        & \multicolumn{1}{l}{return } & \multicolumn{1}{l}{Sortino} & \multicolumn{1}{l}{Sharpe} & \multicolumn{1}{l|}{max DD} \\
    Risky asset 				& 9.16\% &       0.54  &       0.57  & -     0.34  \\
    DRL  						& \textbf{16.42\%} & \textbf{0.98} & \textbf{0.96} & -0.27 \\
    Winner 					& 10.84\% & 0.65  & 0.68  & -0.35  \\
    Loser  					& 7.04\% & 0.78  & 0.76  & \textbf{-0.15}   \\
    DRL no context 				& 6.87\% & 0.44  & 0.47  & -0.34  \\
    Markowitz					& -0.07\% & -0.00  & -0.00  & -0.41  \\
    \bottomrule
    \end{tabular}
}
\end{table}%

\begin{table}[htbp]
  \centering
  \caption{Hyper parameters used}
\resizebox{\linewidth} {!} {
    \begin{tabular}{|l|l|l|}
    \toprule
    hyper-parameters & value  & description \\
    \midrule
    batch size & \multicolumn{1}{r|}{50} & mini-batch size during training \\
    \midrule
    regularization  & \multicolumn{1}{r|}{1e-8}   & $L_2$ regularization coefficient \\
    coefficient & & applied to network training \\
    \midrule
    learning rate & \multicolumn{1}{r|}{0.01} & Step size parameter in Adam \\
    \midrule
    standard deviation  & \multicolumn{1}{r|}{20 days} & period for standard deviation \\
    period & &  in asset states \\
    \midrule
    commission & \multicolumn{1}{r|}{30 bps} & commission rate  \\
    \midrule
    stride & \multicolumn{1}{r|}{2,1} & stride in convolution networks \\
    \midrule
    conv  number 1& \multicolumn{1}{r|}{5,10 } & number of convolutions in \\
    & & sub-network 1\\
    \midrule
    conv  number 2& \multicolumn{1}{r|}{2} & number of convolutions in  \\
    & & sub-network 2\\
    \midrule
    lag period 1 & \multicolumn{1}{r|}{$\! \left[60, \! 20, \!4,\!3,\!2,\!1,\!0 \right] \!$ } & lag period for asset states\\
    \midrule
    lag period 2 & \multicolumn{1}{r|}{$\! \left[ 60,\! 20, \!4,\!3,\!2,\!1,\!0 \right] \! $} & lag period for contextual states\\
    \midrule
     noise & \multicolumn{1}{r|}{0.002} & adversarial Gaussian standard \\
     & &  deviation \\
    \midrule
     max iterations {\large{$^*$}} & \multicolumn{1}{r|}{500} & maximum number of iterations \\
    \midrule
     early stop & \multicolumn{1}{r|}{50} & early stop criterion \\
      iterations {\large{$^*$}}  & & \\
    \midrule
     random seed & \multicolumn{1}{r|}{12345} & random seed \\
    \bottomrule
    \end{tabular}%
}
  \label{tab:hyperparam}%
\end{table}%
\noindent {\large{$^*$}}: 
{\small {the number of iterations is  at maximum 500 provided we do not stop because of early stop detection. We do early stop if on the train set, there is no improvement over the last 50 iterations.}  }

\Section{Conclusion}
In this paper, we address the challenging task of learning in a noisy and self adapting environment with sequential, non-stationary and non-homogeneous observations for a bot and more specifically for a trading bot. Our approach is based on deep reinforcement learning using contextual information thanks to a second sub-network. We also show that the additional constraint of a delayed action following observations has a substantial impact that should not be overlooked. We introduce the novel concept of walk forward to test the robustness of the deep RL model. This is very important for regime changing environment that cannot be evaluated with a simple train validation test procedure, neither a $k$-fold cross validation that ignores the strong chronological feature of observations. 

For our trading bots, we take not only past performances of portfolio strategies over different rolling period, but also standard deviation to provide predictive variables for regime changes. Augmented states with contextual information make a big difference in the model and help the bot learning more efficiently in a noisy environment. On experiment, contextual based approach over-performs baseline methods like Markowitz or naive follow the winner and follow the loser. Last but not least, it is quite important to fine tune the numerous hyper-parameters of the contextual based DRL model, namely the various lags (lags period for the sub network fed by portfolio strategies past returns, lags period for common contextual features referred to as the common features in the paper), standard deviation period, learning rate, etc... 

Despite the efficiency of contextual based DRL models, there is room for improvement. Other information like news could be incorporated to continue increasing model performance. For large stocks, like tech stocks, sentiment information based on social media activity could also be relevant.

\bibliography{main}

\end{document}